\documentclass{article}
\usepackage{amsmath}
\usepackage{amssymb}
\usepackage{amsfonts}
\usepackage{sw20bams}
\usepackage{rotating}

\newtheorem{theorem}{Theorem}

\newtheorem{conclusion}[theorem]{Conclusion}

\newtheorem{definition}[theorem]{Definition}
\newtheorem{example}[theorem]{Example}

\input{tcilatex}
\numberwithin{theorem}{section}

\begin{document}

\title{Predicting Suicide Attacks: A Fuzzy Soft Set Approach}
\author{Athar Kharal\thanks{%
Corresponding author.\newline
Phone: +92 333 6261309} \thanks{%
A version of this manuscript has also been submitted to arXiv.org on 28 July
2010.}\bigskip  \\
National University of Sciences and Technology (NUST),\\
Islamabad, PAKISTAN\bigskip \\
\textit{atharkharal@gmail.com}}
\date{ }
\maketitle

\begin{abstract}
This paper models a decision support system to predict the occurance of
suicide attack in a given collection of cities. The system comprises two
parts. First part analyzes and identifies the factors which affect the
prediction. Admitting incomplete information and use of linguistic terms by
experts, as two characteristic features of this peculiar prediction problem
we exploit the Theory of Fuzzy Soft Sets. Hence the Part 2 of the model is
an algorithm vz. FSP which takes the assessment of factors given in Part 1
as its input and produces a possibility profile of cities likely to receive
the accident. The algorithm is of $O\left( 2^{n}\right) $ complexity. It has
been illustrated by an example solved in detail. Simulation results for the
algorithm have been presented which give insight into the strengths and
weaknesses of FSP. Three different decision making measures have been
simulated and compared in our discussion.\bigskip 

\end{abstract}

\section{\textbf{\protect\bigskip Introduction}}

As this paper was in final stage of preparation, on July 2, 2010 two
simultaneous suicide attacks took place in the Pakistani cosmopolitan city
of Lahore. The attacks left 50 innocent people killed instantly and another
200 severely injured. In total, since 9/11 some 247 suicide attacks took
place in Pakistan killing more than 3000 people. Suicide attacks, when
people themselves are weapons for destruction, have risen globally from an
average of about 3 per year in the 1980s to 40 a year in 2002, almost 50 a
year in 2005 and more than 200 by 2009-10. Globally three fourth of all
suicide bombings have happened since the terror attacks in the USA on
September 11, 2001. Altogether in 16 countries over 300 suicide attacks have
resulted in more than 5,300 people killed and over 10,000 wounded.

In the typical modus operandi of suicide attacks, the terrorist wears an
explosive belt on the body or carries it in a suitcase, sports bag, or
rucksack. Such charges are usually stuffed with about 3--15 kg of TNT or
homemade explosives, together with small chunks of iron or a large quantity
of nails packed around explosives, in order to maximize casualties and loss
of limbs. The detonator of a simple design allows the terrorist to activate
the explosives even when under pressure.

Besides being extremely devastating in terms of casualties, physical damage
and psychological impact such attacks are also highly unpredictable. Many
factors contribute to this high unpredictability. The suicide attackers are
extremely motivated either by ideological and/or religious impulses. An 
\textit{overt attack} i.e. an attack clearly recognizable by law enforcement
agencies and other responding agencies is seldom the choice of attackers. In
almost all cases a \textit{covert attack} strategy where a carefully
disguised dispersal means with no intentional indication, nor a threat or
warning being issued is chosen. Until getting to the specific spot where a
suicide attacker will operate the explosives, only a highly trained eye
would notice anything suspicious. Thus the absence of sufficient empirical
data about suicide attacks presents a peculiar decision making problem
characterized by information which is incomplete, approximate and vague.

As counter-terrorism policies and strategies are being developed across the
globe, prediction of suicide attacks forms an important contribution towards
this end. Besides the incomplete information, predicting suicide attacks is
typically subjective, because it uses the value judgments of experts,
qualitative variables and variables of relative quantification. These
elements make this effort extremely uncertain and prevent the tools and
procedures based on traditional methods from being fully applied. The
decision making environment typically encountered while attempting some
prediction of suicide attacks involves imprecise data and their solution
involves the use of mathematical principles based on uncertainty and
imprecision. Some of these problems are essentially humanistic and thus
subjective in nature, while others are objective, yet they are firmly
embedded in an imprecise environment. Linguistic terms are used by
counter-terror experts and officials but linguistic terms do not hold exact
meaning and may be understood differently by different people. The
boundaries of a given term are rather subjective, and may also depend on the
situation. Linguistic terms therefore cannot be expressed by ordinary set
theory; rather, each decision factor is associated with some linguistic
term. The thought process involved in the act of decision making in a
scenario as defined above is a complex array of streaming possibilities in
which a person selects or discards information made available from diverse
sources. In doing so one is led by a meaningful analysis of available
information and optimal selection out of several apparently equi-efficient
decisions. Information is available on the basis of some criteria and the
criterion-values are not sharply defined but are vague, incomplete and
approximate.

Soft sets introduced by D. Molodtsov \cite{sfst99mol}, is state-of-the-art
approach to deal with incomplete knowledge within information systems.
However, in practical applications of soft sets, the situation is generally
more complex and calls for attributes having non-sharp or fuzzy boundaries,
as is the case with prediction of suicide attacks. That is why many authors
have contributed towards the fuzzification of the notion of soft set e.g. 
\cite%
{aha09-OnFzSftSets,kha09-MpFzSftClasses,kha10-MpSftClasses,kha09-IFzSftStFinanDiag,sfst01maj,sfst07roy,sfst07yan}%
. Fuzzy soft sets is an attractive extension of soft sets, enriching the
latter with extra features to represent uncertainty and vagueness on top of
incompleteness. It is now an area of active research in decision making \cite%
{kha10-Sft approx Decision Mkng} process under vague and approximate
conditions.

In this paper we present a fuzzy soft set based decision making model for
terror attack prediction for a given collection of cities. The paper is
arranged as follows: Section \ref{ModlConst} gives a detailed discussion of
the factors considered important by human analysts for making prediction of
such attacks. After clearly delineating the factors in subsection \ref%
{FactorsOfDescision} to be considered Section \ref{algo} first develops a
general algorithm using fuzzy soft sets for the prediction. This section
besides explaining the working of algorithm and making notes about the
computer implementation, presents a detailed illustrative example. Section %
\ref{sec-simResults} exhibits the results of simulation results and provides
different insights into FSP. Section \ref{sec-discuss} discusses different
aspects of the model; its strengths and weaknesses besides suggesting
possible future enhancements.

\section{\protect\bigskip \label{ModlConst}\textbf{Model Construction}}

Traditional wisdom uses a \textquotedblleft low risk / high
consequence\textquotedblright\ strategy for taking security measures. But
experience has shown it is not the only and sometimes not even the best
counter terror strategy. Scarce resources can be better exploited if a way
is found to predict threats according to likelihood, and not just according
to the severity of consequences.

Explicitly, a model which attempts to predict suicide attacks should cater
for following aspects:

\begin{enumerate}
\item Estimate the \textquotedblleft attractiveness\textquotedblright\ of
specific cities to terrorist organizations,

\item Identify vulnerabilities to build a portfolio of possible attack
locations, and

\item Enable decision makers to concentrate resources on most probable
targets.
\end{enumerate}

Construction of a decision making model under the conditions peculiar to
suicide attacks should cater both the aspects of information incompleteness
and expert's usage of linguistic variables. Linguistic variables are best
dealt with Fuzzy Set Theory of Zadeh \cite{zadeh65} and information
incompleteness is handled by Soft Set Theory of Molodtsov \cite{sfst99mol}.
Both these aspects have been combined in theory of Fuzzy Soft Sets by Maji,
Biswas and Roy \cite{sfst01maj}.

We, therefore, proceed first to identify the main factors which in expert
opinion may contribute towards the selection of target city by suicide
attackers. In the second phase we develop a selection algorithm which takes
advantage of the in-built mechanism for quantization of value judgements of
counter-terror experts.

\subsection{\label{FactorsOfDescision}Potential Factors for Decision Making}

Understanding of the terrorist's target-selection criterion would help us to
identify the potential terror attack locations. Therefore, our assessment of
which targets are most attractive must be made from the point of view of the
terrorist. We enlist following factors to be considered for making a
decision:

\begin{enumerate}
\item \textbf{Maximize the number of casualties:} Their main aim is mass
killing and destruction, without regard for their own life or the lives of
others. Thus the cities with high population density are more prone to such
attacks.

\item \textbf{Damage to the civil/military infrastructure:} As the
terrorists aim to weaken the civil/military infrastructure or at least they
attempt to create an impression that such infrastructures have been
subjugated by and large, cities with high concentration of government
structures, stand more possibility of attack. To cause such damage of
infrastructure, an all alone suicide attack, where a single terrorists
explodes himself among masses, is seldom the choice. Instead a variant of
suicide attack known as Vehicle-borne improvised explosive device (VBIED) is
employed. Typically it is based on a car containing several dozen kilograms
or several hundred pounds of explosives (mostly ANFO= ammonium nitrate,
diesel oil plus a small booster charge). Truck bombs carrying a larger load
of explosives, notably vehicles with up to $8t$ of explosives on board have
already been detonated. The vehicle is parked near an area with increased
population density (marketplace, shopping center, office building) or driven
at high speed against the target (e.g. checkpoint) and detonated either by
remote control or by a suicide driver on board. The type of vehicle used
determines largely the number of casualties resulting from the detonation
(typically several dozen).

\item \textbf{Economic damage potential:} A city where an attack may result
in high economic damage becomes a good choice for terrorists involved in a
relatively protracted war against established government.

\item \textbf{Maximize print and electronic media coverage:} A city with
thorough and prompt media coverage is yet another attraction for terror
groups. Cities with a lot of movement of high profile personalities is
particularly prone to such attacks.

\item \textbf{Vicinity of the areas having religious fanaticism, supremacist
ideology and/or pronounced anti-government ideology:} A city with
geographical and cultural proximity to areas of extremist ideologies is an
easy target. Geographical proximity provides easy access to urban center and
cultural proximity provides much need camouflage for the suicide attacker.

\item \textbf{Day of week:} Day of week has been observed to be of special
importance. For example a day of religious gathering of another sect
immediately draws the attention of terrorists. Equivalently, in certain
countries people of highest military and civil establishment gather for
weekly religious services and thus become a very attractive target for
suicide bombers.

\item \textbf{Post attack hiding:} A city which either due to its high
population density or due to its urban planning provides easy hiding places
or escape routes is another factor contributing towards the choice of
suicide attacker and his accomplices.

\item \textbf{Safe heaven:} City with access to a relatively closer safe
heaven where development efforts can take place unhindered is very high on
terrorist's priority list.

\item \textbf{Physical difficulty:} It pertains to different physical
obstacles presented to the attacker e.g. security pickets, checkposts, body
search spots etc.

\item \textbf{Psychological impact:} A city or location generally viewed as
the fortress of top leadership and/or establishment (civil/military) carries
a very high potential for psychological impact in case of attack. For
example such an impact will be very high for capital of a country.
\end{enumerate}

It is important to realize that each of the above mentioned factors is a
fuzzy set over the universal set of cities, where membership grade of each
city may be decided either quantitatively as in case of population density
for factor $1$ and/or $7$ or by expert's opinion as in factors $4$ and $10.$
Some of the factors may seem overlapping e.g. $5$ and $6,$ but as the
prediction attempt suffers from vagueness of linguistic terms and incomplete
knowledge both, such an overlap is in fact helpful for passing a judgement
based upon plausible reasoning. Moreover factor $5$ is a relatively
qualitative measure but $6$ is quantitatively measurable. Thus each
complements the other. A similar argument is extended for the other
resembling factors.

Because such a decision making is generally carried out collectively e.g. by
a committee of counter-terror officials, it is very much possible that
either a certain factor remains undecided or no information could be
gathered for it at all. In both cases the factor has to be eliminated thus
modelling the incomplete information aspect of the problem. Secondly, as the
linguistic terms used by experts are fuzzy sets we are naturally driven to
model this decision making problem using fuzzy soft sets. Therefor we first
examine some basic notions from Soft Set Theory and their relevance to our
model.

\section{\protect\bigskip \textbf{Development of Algorithm}}

As mentioned earlier this prediction problem has two peculiar
characteristics:

\begin{enumerate}
\item Use of linguistic terms by counter-terror experts in their subjective
assessments,

\item Incomplete information and hence the presence of approximate
descriptions of different factors of interest.
\end{enumerate}

It is well established now that linguistic vagueness is best tackled by
Fuzzy Set Theory \cite{zadeh65}. On the other hand, Molodtsov \cite%
{sfst99mol} introduced the Theory off Soft Sets to handle the approximate
descriptions. Maji, Biswas and Roy \cite{sfst01maj} combined both these
theories as Fuzzy Soft Sets. Naturally this synergy of views is better
suited to deal with the problems of linguistic vagueness coupled with
approximate descriptions. Thus we intend to develop an algorithm using fuzzy
soft sets. For this we first introduce the basic notions of theory:

\begin{definition}
A pair $(F,A)$ is called a soft set \cite{sfst99mol} over $X$, where $F$ is
a mapping given by $F:A\rightarrow P(X).$\newline
In other words, a soft set over $X$ is a parametrized family of subsets of
the universe $X.$ For $\varepsilon \in A,$ $F(\varepsilon )$ may be
considered as the set of $\varepsilon $-approximate elements of the soft set 
$(F,A)$. Clearly a soft set is not a set in ordinary sense.
\end{definition}

\begin{definition}
\cite{kha10-MpSftClasses} Let $X$ be a universe and $E$ a set of attributes.
Then the pair $\left( X,E\right) ,$ called a soft space, is the collection
of all soft sets on $X$ with attributes from $E$.
\end{definition}

Maji \textit{et al.} defined a fuzzy soft set in the following manner:

\begin{definition}
\cite{sfst01maj} A pair $(\Lambda ,\Sigma )$ is called a fuzzy soft set over 
$X,$ where $\Lambda :\Sigma \rightarrow \widetilde{P}\left( X\right) $ is\ a
mapping, $\widetilde{P}(X)$ being the set of all fuzzy sets of $X.$
\end{definition}

For our model the set of cities is the universal set and the factors of
decision making may be viewed as the set $E.$ As the factors for decision
making are linguistic variables thus $E$ has to be a set of linguistic
variables rendering $\left( X,E\right) $ to be a fuzzy soft space. Fuzzy
soft space is defined as:

\begin{definition}
\cite{kha09-MpFzSftClasses} Let $X$ be an universe and $E$ a set of
attributes. Then the collection of all fuzzy soft sets over $X$ with
attributes from $E$ is called a fuzzy soft class and is denoted as $%
\widetilde{\left( X,E\right) }$.
\end{definition}

\begin{example}
Let $X=\left\{ a,b,c,d,e\right\} $ be the set of cities and $E,$ the set of
decision factors given as: 
\begin{equation*}
\begin{tabular}{lllll}
$\varepsilon _{1}=$ & High count of casualties &  & $\varepsilon _{6}=$ & 
Cultural proximity to terrorists' ethnicity \\ 
$\varepsilon _{2}=$ & Damage to infrastructure &  & $\varepsilon _{7}=$ & 
Pre attack hiding \\ 
$\varepsilon _{3}=$ & Economic damage &  & $\varepsilon _{8}=$ & Post attack
hiding \\ 
$\varepsilon _{4}=$ & Media Coverage &  & $\varepsilon _{9}=$ & Physical
security \\ 
$\varepsilon _{5}=$ & Distance from terrorist strongholds &  & $\varepsilon
_{10}=$ & Psychological impact%
\end{tabular}%
\end{equation*}%
Then the assessment of each parameter for every city, as made by a committee
of experts, can easily be represented in the form of following fuzzy soft
set:%
\begin{equation*}
\left( \Lambda ,\Sigma \right) =\left\{ 
\begin{array}{c}
\varepsilon _{1}=\left\{ a_{0.7},b_{1},c_{1},d_{0.8},e_{1}\right\}
,\varepsilon _{2}=\left\{ a_{1},b_{0.2},c_{0.9},d_{1},e_{0.2}\right\}
,\varepsilon _{3}=\left\{ a_{0.6},b_{0.2},c_{0.1},d_{0.3},e_{0.8}\right\} ,
\\ 
\varepsilon _{4}=\left\{ a_{0.2},b_{0.4},c_{0.6},d_{0.1},e_{0.4}\right\}
,\varepsilon _{5}=\left\{ a_{0.4},b_{0.8},c_{0.7},d_{0.1},e_{0.2}\right\}
,\varepsilon _{6}=\left\{ a_{0.6},b_{0.3},c_{0.7},d_{0.3},e_{0.9}\right\} ,
\\ 
\varepsilon _{7}=\left\{ a_{0.5},b_{0.9},c_{0.3},d_{0.5},e_{0.5}\right\}
,\varepsilon _{8}=\left\{ a_{0.1},b_{1},c_{0.3},d_{0.5},e_{0.9}\right\}
,\varepsilon _{9}=\left\{ a_{0.8},b_{0.2},c_{0.5},d_{1},e_{0.7}\right\} , \\ 
\varepsilon _{10}=\left\{ a_{0.5},b_{0.8},c_{0.3},d_{1},e_{0.4}\right\} 
\end{array}%
\right\} 
\end{equation*}%
\bigskip \bigskip 
\end{example}

For a given pair of cities and a given potential factor of prediction, a
city having higher membership grade is preferred over the other city. Such
pairwise comparisons, if carried out against each decision factor would
yield a list of preferences against each decision factor. List of such
dominations and subjections for each city may be collected in a similar
manner. Heart of the decision making technique is the aggregation of these
lists in a meaningful manner. In other words, an aggregation of both lists
for a given city translates into a decision measure. Thus the technique aims
to exploit comparison of a given city's total dominations vis-a-vis its
subjections and magnifying this quotient by the number of equally graded
attributes.  Using this line of thought we develop algorithm FSP (Fuzzy Soft
Prediction) as follows:

\subsection{\label{algo}Algorithm}

\begin{gather*}
\begin{tabular}{lll}
\textbf{Name} & : & F\textbf{SP} \\ 
\textbf{Input} & : & Set of cities and the factors deemed suitable for
prediction. \\ 
&  &  \\ 
\textbf{Output} & : & Sorted list of cities with most probable location at
the top. \\ 
&  & 
\end{tabular}
\\
\begin{tabular}{|ll|}
\hline
&  \\ 
1. & $X\leftarrow \left\{ \psi _{i}\right\} $ \\ 
&  \\ 
2. & $E\leftarrow \left\{ \varepsilon _{i}\right\} $ \\ 
&  \\ 
3. & $\left( \Lambda ,\Xi \right) \leftarrow \left\{ \varepsilon
_{i}=\Lambda \left( \varepsilon _{i}\right) ~|~\Lambda \left( \varepsilon
_{i}\right) \in \left[ 0,1\right] ^{X}\right\} \in \widetilde{\left[ X,E%
\right] }$ \\ 
&  \\ 
4. & \textbf{FOR} $\left( \psi _{i},\psi _{j}\right) \in X^{2}$ calculate
and store \\ 
5. & $~~~~~~~~~~~~~~~~~~~\rho \left( \psi _{i},\psi _{j}\right) =\left\{
\varepsilon _{i}\in \Xi ~|~\Lambda \left( \psi _{i}\right) \geq \Lambda
\left( \psi _{j}\right) \right\} $ \\ 
6. & $~~~~~~~~~~~~~~~~~~~\chi \left( \psi _{i},\psi _{j}\right) =\left\{
\varepsilon _{i}\in \Xi ~|~\Lambda \left( \psi _{i}\right) \leq \Lambda
\left( \psi _{j}\right) \right\} $ \\ 
7. & \textbf{END FOR} \\ 
&  \\ 
8. & \textbf{FOR} $\psi _{i}\in X$ calculate~and store \\ 
9. & $~~~~~~~~~~~~~~~~~~~\triangledown \left( \psi _{i}\right)
=\dsum\limits_{\psi _{j}\in X}\left\Vert \rho \left( \psi _{i},\psi
_{j}\right) \right\Vert $ \\ 
10. & $~~~~~~~~~~~~~~~~~~~\Delta \left( \psi _{i}\right) =\dsum\limits_{\psi
_{j}\in X}\left\Vert \chi \left( \psi _{i},\psi _{j}\right) \right\Vert $ \\ 
&  \\ 
11. & $~~~~~~~~~~~~~~~~~~~\Omega \left( \psi _{i}\right) =\dsum\limits_{\psi
_{j}\in X}\left\Vert \rho \left( \psi _{i},\psi _{j}\right) \cap \chi \left(
\psi _{i},\psi _{j}\right) \right\Vert $ \\ 
&  \\ 
12. & $~~~~~~~~~~~~~~~~~~~\Gamma _{1}\left( \psi _{i}\right) =\dfrac{%
\triangledown \left( \psi _{i}\right) }{\Delta \left( \psi _{i}\right) }%
\times \Omega \left( \psi _{i}\right) $ \\ 
13. & \textbf{END FOR} \\ 
&  \\ 
14. & $DecisionTable\leftarrow $\textbf{SORT} $\left( \psi _{i},\Gamma
_{1}\left( \psi _{i}\right) \right) $ in descending order of $\Gamma
_{1}\left( \psi _{i}\right) $ \\ 
&  \\ 
15 & \textbf{OUTPUT} $DecisionTable$ \\ 
&  \\ \hline
\end{tabular}%
\end{gather*}

Where $\left\Vert .\right\Vert $ denotes the cardinality of a crisp set.

\subsection{\label{AlgoExplained}Explanation of FSP}

FSP exploits comparison of a given city's total dominations vis-a-vis its
subordinations and magnifying this quotient by the number of equally graded
attributes. For given two cities, domination is an attribute and/or a
decision factor in which one city has a higher grade of membership as
compared to the other one. Similarly, subjection is the factor where one
city has a lesser membership grade than the other one. In FSP, $\rho $ and $%
\chi $ denote the dominations and subjections, respectively. A full listing
of $\rho \left( \psi _{i},\psi _{j}\right) $ and $\chi \left( \psi _{i},\psi
_{j}\right) $ looks like the discernibility matrix of Skowron \cite%
{skowron91,pawlak91bk} in Rough Set Theory.

Some explanation of decision measure $\Gamma _{1}$ is in point here. $\Gamma
_{1}$ is an equities weighted ratio of dominations and subjections of a
city. The role of $\Omega $, total number of attributes having equal
membership grades may be considered less useful for certain situations. In
such case we may compute $\Gamma _{2}$ given as:%
\begin{equation*}
\Gamma _{2}=\triangledown \left( \psi _{i}\right) -\Delta \left( \psi
_{i}\right) .
\end{equation*}%
Clearly $\Gamma _{2}$ is simple measure of differentiation or distance
between the dominations and subjections of a city. Yet another decision
measure is $\Gamma _{3}$ which scales total decision attributes by the
equity factor and is given as:%
\begin{equation*}
\Gamma _{3}=\dfrac{\triangledown \left( \psi _{i}\right) +\Delta \left( \psi
_{i}\right) }{\Omega \left( \psi _{i}\right) }.
\end{equation*}%
In general, $\rho \left( \psi _{i},\psi _{j}\right) +\chi \left( \psi
_{i},\psi _{j}\right) \neq \left\Vert \Xi \right\Vert .$ This makes all
three measures independent of each other and thus providing three different
views of the same decision environment.

\subsection{Computer Implementation of FSP}

For computer implementation a class named FSS is introduced which specifies
the notion of fuzzy soft sets. The basic data structure of FSP is a two
dimensional array. This array is constructed in Step $3$ from the available
sets in Step $1,2$ and is a member of the class FSS. First part of Step $4$
(lines $4-7$) determines the attribute set for each pair of cities where
first city dominates others and second part compares for dominated-by pairs.
Thus $\rho \left( \psi _{i},\psi _{j}\right) $ is the domination attribute
set of $\psi _{i}$ and $\chi \left( \psi _{i},\psi _{j}\right) $ is the
dominated-by attribute set of cities $\psi _{i}$ and $\psi _{j}.$ Step $8$
(lines $8-11$) collects total number of dominations, subjections and
equities for each city $\psi _{i}.$ Line $12$ computes the decision measure $%
\Gamma _{1}.$

For further explaining the algorithm we give an illustrative example in the
following:

\begin{example}
Since 9/11 in USA, there have been some 247 suicide attacks in Pakistan
killing about 3000 innocent lives.. On 12th March 2010, Lahore, one of the
five major cities of Pakistan, encountered a spate of nine bombings in a
single day killing about 100 people. Out of the nine incidents $4$ were
suicide attacks. In this example we apply our model of factors and the
algorithm FSP to a multi-observer based committee of experts' opinion about
the next possible suicide attack in Pakistan. We specifically consider $4$
provincial capitals and the national capital. Figure \ref{PakMap} shows map
of Pakistan with all the major cities. 
\begin{equation*}
\begin{tabular}{lll}
$%
\begin{tabular}{ll}
$\psi _{1}=$ & Karachi \\ 
&  \\ 
$\psi _{2}=$ & Lahore \\ 
&  \\ 
$\psi _{3}=$ & Quetta \\ 
&  \\ 
$\psi _{4}=$ & Islamabad \\ 
&  \\ 
$\psi _{5}=$ & Peshawar%
\end{tabular}%
$ & ~~~~~~~~~~~~~ & $%
\begin{tabular}{l}
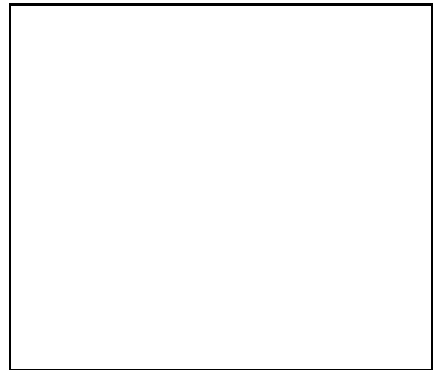
\FRAME{itbpFUX}{2.2044in}{1.9112in}{0in}{\Qcb{Major cities of Pakistan}}{%
\Qlb{PakMap}}{pakistanmap.tif}{\special{language "Scientific Word";type
"GRAPHIC";maintain-aspect-ratio TRUE;display "USEDEF";valid_file "F";width
2.2044in;height 1.9112in;depth 0in;original-width 7.7816in;original-height
6.7395in;cropleft "0";croptop "1";cropright "1";cropbottom "0";filename
'PakistanMap.TIF';file-properties "XNPEU";}}%
\end{tabular}%
$%
\end{tabular}%
\end{equation*}%
Decision parameters as explicated in Section \ref{FactorsOfDescision} have
been symbolized as follows:%
\begin{equation*}
\begin{tabular}{lllll}
$\varepsilon _{1}=$ & High count of casualties &  & $\varepsilon _{6}=$ & 
Cultural proximity to terrorists' ethnicity \\ 
&  &  &  &  \\ 
$\varepsilon _{2}=$ & Damage to infrastructure &  & $\varepsilon _{7}=$ & 
Pre attack hiding \\ 
&  &  &  &  \\ 
$\varepsilon _{3}=$ & Economic damage &  & $\varepsilon _{8}=$ & Post attack
hiding \\ 
&  &  &  &  \\ 
$\varepsilon _{4}=$ & Media Coverage &  & $\varepsilon _{9}=$ & Physical
security. \\ 
&  &  &  &  \\ 
$\varepsilon _{5}=$ & Distance from terrorist strongholds &  & $\varepsilon
_{10}=$ & Psychological impact%
\end{tabular}%
\end{equation*}%
Then the assessment of each parameter for every city, as made by a committee
of experts, can easily be represented in the form of following fuzzy soft
set: 
\begin{equation*}
\begin{tabular}{|l|llllllllll|}
\hline
\multicolumn{1}{|l|}{} & $\mathbf{\varepsilon }_{1}$ & $\mathbf{\varepsilon }%
_{2}$ & $\mathbf{\varepsilon }_{3}$ & $\mathbf{\varepsilon }_{4}$ & $\mathbf{%
\varepsilon }_{5}$ & $\mathbf{\varepsilon }_{6}$ & $\mathbf{\varepsilon }_{7}
$ & $\mathbf{\varepsilon }_{8}$ & $\mathbf{\varepsilon }_{9}$ & $\mathbf{%
\varepsilon }_{10}$ \\ \hline
&  &  &  &  &  &  &  &  &  &  \\ 
$\mathbf{\psi }_{1}$ & $0.7$ & $1.0$ & $0.6$ & $0.2$ & $0.4$ & $0.6$ & $0.5$
& $0.1$ & $0.8$ & $0.5$ \\ 
&  &  &  &  &  &  &  &  &  &  \\ 
$\mathbf{\psi }_{2}$ & $1.0$ & $0.2$ & $0.2$ & $0.4$ & $0.8$ & $0.3$ & $0.9$
& $1.0$ & $0.2$ & $0.8$ \\ 
&  &  &  &  &  &  &  &  &  &  \\ 
$\mathbf{\psi }_{3}$ & $1.0$ & $0.9$ & $0.1$ & $0.6$ & $0.7$ & $0.7$ & $0.3$
& $0.3$ & $0.5$ & $0.3$ \\ 
&  &  &  &  &  &  &  &  &  &  \\ 
$\mathbf{\psi }_{4}$ & $0.8$ & $1.0$ & $0.3$ & $0.1$ & $0.1$ & $0.3$ & $0.5$
& $0.5$ & $1.0$ & $1.0$ \\ 
&  &  &  &  &  &  &  &  &  &  \\ 
$\mathbf{\psi }_{5}$ & $1.0$ & $0.2$ & $0.8$ & $0.4$ & $0.2$ & $0.9$ & $0.5$
& $0.9$ & $0.7$ & $0.4$ \\ 
&  &  &  &  &  &  &  &  &  &  \\ \hline
\end{tabular}%
\end{equation*}%
Tables \ref{TableOfRho} and \ref{TableOfChi} on next page, give the
dominations and subjections of each city. 
\begin{sidewaystable}\centering%
$%
\begin{tabular}{|c|ccccc|}
\hline
$\mathbf{\rho }\left( \psi _{i},\psi _{j}\right) $ & $\mathbf{\psi }_{1}$ & $%
\mathbf{\psi }_{2}$ & $\mathbf{\psi }_{3}$ & $\mathbf{\psi }_{4}$ & $\mathbf{%
\psi }_{5}$ \\ \hline
&  &  &  &  &  \\ 
$\mathbf{\psi }_{1}$ & $E$ & $\left\{ \varepsilon _{2},\varepsilon
_{3},\varepsilon _{6},\varepsilon _{9}\right\} $ & $\left\{ \varepsilon
_{2},\varepsilon _{3},\varepsilon _{7},\varepsilon _{9},\varepsilon
_{10}\right\} $ & $\left\{ \varepsilon _{2},\varepsilon _{3},\varepsilon
_{4},\varepsilon _{5},\varepsilon _{6},\varepsilon _{7}\right\} $ & $\left\{
\varepsilon _{2},\varepsilon _{5},\varepsilon _{7},\varepsilon
_{9},\varepsilon _{10}\right\} $ \\ 
&  &  &  &  &  \\ 
$\mathbf{\psi }_{2}$ & $\left\{ \varepsilon _{1},\varepsilon
_{4},\varepsilon _{5},\varepsilon _{7},\varepsilon _{8},\varepsilon
_{10}\right\} $ & $E$ & $\left\{ \varepsilon _{1},\varepsilon
_{3},\varepsilon _{5},\varepsilon _{7},\varepsilon _{8},\varepsilon
_{10}\right\} $ & $\left\{ \varepsilon _{1},\varepsilon _{4},\varepsilon
_{5},\varepsilon _{6},\varepsilon _{7},\varepsilon _{8}\right\} $ & $\left\{
\varepsilon _{1},\varepsilon _{2},\varepsilon _{4},\varepsilon
_{5},\varepsilon _{7},\varepsilon _{8},\varepsilon _{10}\right\} $ \\ 
&  &  &  &  &  \\ 
$\mathbf{\psi }_{3}$ & $\left\{ \varepsilon _{1},\varepsilon
_{4},\varepsilon _{5},\varepsilon _{6},\varepsilon _{8}\right\} $ & $\left\{
\varepsilon _{1},\varepsilon _{2},\varepsilon _{4},\varepsilon
_{6},\varepsilon _{9}\right\} $ & $E$ & $\left\{ \varepsilon
_{1},\varepsilon _{4},\varepsilon _{5},\varepsilon _{6}\right\} $ & $\left\{
\varepsilon _{1},\varepsilon _{2},\varepsilon _{4},\varepsilon _{5}\right\} $
\\ 
&  &  &  &  &  \\ 
$\mathbf{\psi }_{4}$ & $\left\{ \varepsilon _{1},\varepsilon
_{2},\varepsilon _{7},\varepsilon _{8},\varepsilon _{9},\varepsilon
_{10}\right\} $ & $\left\{ \varepsilon _{2},\varepsilon _{3},\varepsilon
_{6},\varepsilon _{9},\varepsilon _{10}\right\} $ & $\left\{ \varepsilon
_{2},\varepsilon _{3},\varepsilon _{7},\varepsilon _{8},\varepsilon
_{9},\varepsilon _{10}\right\} $ & $E$ & $\left\{ \varepsilon
_{2},\varepsilon _{7},\varepsilon _{9},\varepsilon _{10}\right\} $ \\ 
&  &  &  &  &  \\ 
$\mathbf{\psi }_{5}$ & $\left\{ \varepsilon _{1},\varepsilon
_{3},\varepsilon _{4},\varepsilon _{6},\varepsilon _{7},\varepsilon
_{8}\right\} $ & $\left\{ \varepsilon _{1},\varepsilon _{2},\varepsilon
_{3},\varepsilon _{4},\varepsilon _{6},\varepsilon _{9}\right\} $ & $\left\{
\varepsilon _{1},\varepsilon _{3},\varepsilon _{6},\varepsilon
_{7},\varepsilon _{8},\varepsilon _{9},\varepsilon _{10}\right\} $ & $%
\left\{ \varepsilon _{1},\varepsilon _{3},\varepsilon _{4},\varepsilon
_{5},\varepsilon _{6},\varepsilon _{7},\varepsilon _{8}\right\} $ & $E$ \\ 
&  &  &  &  &  \\ \hline
\end{tabular}%
$\caption{Table of dominations}\label{TableOfRho}\bigskip \bigskip \bigskip
\bigskip $%
\begin{tabular}{|cccccc|}
\hline
\multicolumn{1}{|c|}{$\mathbf{\chi }\left( \psi _{i},\psi _{j}\right) $} & $%
\mathbf{\psi }_{1}$ & $\mathbf{\psi }_{2}$ & $\mathbf{\psi }_{3}$ & $\mathbf{%
\psi }_{4}$ & $\mathbf{\psi }_{5}$ \\ \hline
\multicolumn{1}{|c|}{} & \multicolumn{1}{|c}{} &  &  &  &  \\ 
$\mathbf{\psi }_{1}$ & \multicolumn{1}{|c}{$E$} & $\left\{ \varepsilon
_{1},\varepsilon _{4},\varepsilon _{5},\varepsilon _{7},\varepsilon
_{8},\varepsilon _{10}\right\} $ & $\left\{ \varepsilon _{1},\varepsilon
_{4},\varepsilon _{5},\varepsilon _{6},\varepsilon _{8}\right\} $ & $\left\{
\varepsilon _{1},\varepsilon _{2},\varepsilon _{7},\varepsilon
_{8},\varepsilon _{9},\varepsilon _{10}\right\} $ & $\left\{ \varepsilon
_{1},\varepsilon _{3},\varepsilon _{4},\varepsilon _{6},\varepsilon
_{7},\varepsilon _{8}\right\} $ \\ 
& \multicolumn{1}{|c}{} &  &  &  &  \\ 
\multicolumn{1}{|c|}{$\mathbf{\psi }_{2}$} & \multicolumn{1}{|c}{$\left\{
\varepsilon _{2},\varepsilon _{3},\varepsilon _{6},\varepsilon _{9}\right\} $%
} & $E$ & $\left\{ \varepsilon _{1},\varepsilon _{2},\varepsilon
_{4},\varepsilon _{6},\varepsilon _{9}\right\} $ & $\left\{ \varepsilon
_{2},\varepsilon _{3},\varepsilon _{6},\varepsilon _{9},\varepsilon
_{10}\right\} $ & $\left\{ \varepsilon _{1},\varepsilon _{2},\varepsilon
_{3},\varepsilon _{4},\varepsilon _{6},\varepsilon _{9}\right\} $ \\ 
& \multicolumn{1}{|c}{} &  &  &  &  \\ 
\multicolumn{1}{|c|}{$\mathbf{\psi }_{3}$} & \multicolumn{1}{|c}{$\left\{
\varepsilon _{2},\varepsilon _{3},\varepsilon _{7},\varepsilon
_{9},\varepsilon _{10}\right\} $} & $\left\{ \varepsilon _{1},\varepsilon
_{3},\varepsilon _{5},\varepsilon _{7},\varepsilon _{8},\varepsilon
_{10}\right\} $ & $E$ & $\left\{ \varepsilon _{2},\varepsilon
_{3},\varepsilon _{7},\varepsilon _{8},\varepsilon _{9},\varepsilon
_{10}\right\} $ & $\left\{ \varepsilon _{1},\varepsilon _{3},\varepsilon
_{6},\varepsilon _{7},\varepsilon _{8},\varepsilon _{9},\varepsilon
_{10}\right\} $ \\ 
& \multicolumn{1}{|c}{} &  &  &  &  \\ 
\multicolumn{1}{|c|}{$\mathbf{\psi }_{4}$} & \multicolumn{1}{|c}{$\left\{
\varepsilon _{2},\varepsilon _{3},\varepsilon _{4},\varepsilon
_{5},\varepsilon _{6},\varepsilon _{7}\right\} $} & $\left\{ \varepsilon
_{1},\varepsilon _{4},\varepsilon _{5},\varepsilon _{6},\varepsilon
_{7},\varepsilon _{8}\right\} $ & $\left\{ \varepsilon _{1},\varepsilon
_{4},\varepsilon _{5},\varepsilon _{6}\right\} $ & $E$ & $\left\{
\varepsilon _{1},\varepsilon _{3},\varepsilon _{4},\varepsilon
_{5},\varepsilon _{6},\varepsilon _{7},\varepsilon _{8}\right\} $ \\ 
& \multicolumn{1}{|c}{} &  &  &  &  \\ 
$\mathbf{\psi }_{5}$ & \multicolumn{1}{|c}{$\left\{ \varepsilon
_{2},\varepsilon _{5},\varepsilon _{7},\varepsilon _{9},\varepsilon
_{10}\right\} $} & $\left\{ \varepsilon _{1},\varepsilon _{2},\varepsilon
_{4},\varepsilon _{5},\varepsilon _{7},\varepsilon _{8},\varepsilon
_{10}\right\} $ & $\left\{ \varepsilon _{1},\varepsilon _{2},\varepsilon
_{4},\varepsilon _{5}\right\} $ & $\left\{ \varepsilon _{2},\varepsilon
_{7},\varepsilon _{9},\varepsilon _{10}\right\} $ & $E$ \\ 
\multicolumn{1}{|c|}{} & \multicolumn{1}{|c}{} &  &  &  &  \\ \hline
\end{tabular}%
$\caption{Table of subjections}\label{TableOfChi}%
\end{sidewaystable}
This permits to calculate cumulative domination, subjection and equity for
each particular city as follows:%
\begin{equation*}
\begin{tabular}{lllll}
$\triangledown \left( \psi _{1}\right) =30,$ &  & $\Delta \left( \psi
_{1}\right) =33,$ &  & $\Omega \left( \psi _{1}\right) =13$ \\ 
&  &  &  &  \\ 
$\triangledown \left( \psi _{2}\right) =35,$ &  & $\Delta \left( \psi
_{2}\right) =30,$ &  & $\Omega \left( \psi _{2}\right) =15$ \\ 
&  &  &  &  \\ 
$\triangledown \left( \psi _{3}\right) =28,$ &  & $\Delta \left( \psi
_{3}\right) =34,$ &  & $\Omega \left( \psi _{3}\right) =12$ \\ 
&  &  &  &  \\ 
$\triangledown \left( \psi _{4}\right) =31,$ &  & $\Delta \left( \psi
_{4}\right) =33,$ &  & $\Omega \left( \psi _{4}\right) =14$ \\ 
&  &  &  &  \\ 
$\triangledown \left( \psi _{5}\right) =36,$ &  & $\Delta \left( \psi
_{5}\right) =30,$ &  & $\Omega \left( \psi _{5}\right) =16$ \\ 
&  &  &  & 
\end{tabular}%
\end{equation*}%
Calculating $\Gamma _{1},$ $\Gamma _{2}$, $\Gamma _{3}$ and plotting the
three possibility distributions we get%
\begin{equation*}
\begin{tabular}{l|ccccc}
& $\mathbf{\Gamma }_{1}$ &  & $\mathbf{\Gamma }_{2}$ &  & $\mathbf{\Gamma }%
_{3}$ \\ \hline
&  &  &  &  &  \\ 
$\mathbf{\psi }_{1}$ & \multicolumn{1}{|l}{$11.8$} & \multicolumn{1}{l}{} & 
\multicolumn{1}{l}{$-3$} & \multicolumn{1}{l}{} & \multicolumn{1}{l}{$4.9$}
\\ 
&  &  &  &  &  \\ 
$\mathbf{\psi }_{2}$ & \multicolumn{1}{|l}{$17.5$} & \multicolumn{1}{l}{} & 
\multicolumn{1}{l}{$~~5$} & \multicolumn{1}{l}{} & \multicolumn{1}{l}{$4.33$}
\\ 
&  &  &  &  &  \\ 
$\mathbf{\psi }_{3}$ & \multicolumn{1}{|l}{$9.9$} & \multicolumn{1}{l}{} & 
\multicolumn{1}{l}{$-6$} & \multicolumn{1}{l}{} & \multicolumn{1}{l}{$5.17$}
\\ 
&  &  &  &  &  \\ 
$\mathbf{\psi }_{4}$ & \multicolumn{1}{|l}{$13.2$} & \multicolumn{1}{l}{} & 
\multicolumn{1}{l}{$-2$} & \multicolumn{1}{l}{} & \multicolumn{1}{l}{$4.57$}
\\ 
&  &  &  &  &  \\ 
$\mathbf{\psi }_{5}$ & \multicolumn{1}{|l}{$19.2$} & \multicolumn{1}{l}{} & 
\multicolumn{1}{l}{$~~6$} & \multicolumn{1}{l}{} & \multicolumn{1}{l}{$4.12$}
\\ 
&  &  &  &  & 
\end{tabular}%
,~~\FRAME{itbpF}{2.8253in}{2.0064in}{0.8043in}{}{}{decisiondistribution.tif}{%
\special{language "Scientific Word";type "GRAPHIC";display "PICT";valid_file
"F";width 2.8253in;height 2.0064in;depth 0.8043in;original-width
5.0211in;original-height 3.0104in;cropleft "0";croptop "1";cropright
"1";cropbottom "0";filename 'DecisionDistribution.TIF';file-properties
"XNPEU";}}
\end{equation*}%
Sorting decision table under $\Gamma _{1}$ we have%
\begin{equation*}
\begin{tabular}{|l|ccccc|}
\hline
& $\mathbf{\Gamma }_{1}$ &  & $\mathbf{\Gamma }_{2}$ &  & $\mathbf{\Gamma }%
_{3}$ \\ \hline
&  &  &  &  &  \\ 
$\mathbf{\psi }_{5}$ & $19.2$ &  & $~~6$ &  & $4.12$ \\ 
&  &  &  &  &  \\ 
$\mathbf{\psi }_{2}$ & $17.5$ &  & $~~5$ &  & $4.33$ \\ 
&  &  &  &  &  \\ 
$\mathbf{\psi }_{4}$ & $13.2$ &  & $-2$ &  & $4.57$ \\ 
&  &  &  &  &  \\ 
$\mathbf{\psi }_{1}$ & $11.8$ &  & $-3$ &  & $4.9$ \\ 
&  &  &  &  &  \\ 
$\mathbf{\psi }_{3}$ & $9.9$ &  & $-6$ &  & $5.17$ \\ \hline
\end{tabular}%
\end{equation*}%
Thus the city of Peshawar $\left( \psi _{5}\right) $ is the most likely next
target of a suicide attack according to the given expert assessment.
\end{example}

\section{\label{sec-simResults}\protect\bigskip \textbf{Simulation Results}}

It is imperative to estimate the bias (if any) of FSP. Using Borland's C++
compiler on an Intel Pentium dual core machine under Windows operating
system, a sample run of $1000$ scenarios was simulated for $10$ cities with $%
20$ factors of decision. The frequency of each city being most probable
target is counted as:%
\begin{equation*}
\begin{tabular}{|c|cccccccccc|}
\hline
& $\mathbf{\psi }_{1}$ & $\mathbf{\psi }_{2}$ & $\mathbf{\psi }_{3}$ & $%
\mathbf{\psi }_{4}$ & $\mathbf{\psi }_{5}$ & $\mathbf{\psi }_{6}$ & $\mathbf{%
\psi }_{7}$ & $\mathbf{\psi }_{8}$ & $\mathbf{\psi }_{9}$ & $\mathbf{\psi }%
_{10}$ \\ \hline
&  &  &  &  &  &  &  &  &  &  \\ 
$\mathbf{\Gamma }_{1}$ & $103$ & $101$ & $95$ & $83$ & $98$ & $103$ & $108$
& $98$ & $107$ & $118$ \\ 
&  &  &  &  &  &  &  &  &  &  \\ 
$\mathbf{\Gamma }_{2}$ & $102$ & $100$ & $96$ & $82$ & $101$ & $91$ & $123$
& $118$ & $109$ & $109$ \\ 
&  &  &  &  &  &  &  &  &  &  \\ 
$\mathbf{\Gamma }_{3}$ & $114$ & $122$ & $105$ & $131$ & $119$ & $117$ & $146
$ & $128$ & $117$ & $130$ \\ 
&  &  &  &  &  &  &  &  &  &  \\ \hline
\end{tabular}%
\end{equation*}%
Histogram of decision experiments is shown below exhibiting a non-biased
evenly distributed distribution for all cities:

\FRAME{dtbpF}{5.0756in}{3.0545in}{0pt}{}{}{mostprobablecityfrequency.tif}{%
\special{language "Scientific Word";type "GRAPHIC";maintain-aspect-ratio
TRUE;display "USEDEF";valid_file "F";width 5.0756in;height 3.0545in;depth
0pt;original-width 5.0211in;original-height 3.0104in;cropleft "0";croptop
"1";cropright "1";cropbottom "0";filename
'MostProbableCityFrequency.TIF';file-properties "XNPEU";}}

For certain scenarios higher number of ties may be considered weakness of a
decision measure. This helps to choose between $\Gamma _{1},\Gamma _{2}$ and 
$\Gamma _{3}.$ Following is a count of ties in a sample run of the
simulation:%
\begin{equation*}
\begin{tabular}{|ccccc|}
\hline
$\mathbf{\Gamma }_{1}$ &  & $\mathbf{\Gamma }_{2}$ &  & $\mathbf{\Gamma }%
_{3} $ \\ \hline
&  &  &  &  \\ 
$246$ &  & $440$ &  & $2129$ \\ 
&  &  &  &  \\ \hline
\end{tabular}%
\end{equation*}

\FRAME{dtbpFU}{5.0756in}{3.0545in}{0pt}{\Qcb{Number of ties in $\Gamma _{1},$
$\Gamma _{2}$ and $\Gamma _{3}.$}}{}{noofgammaties.tif}{\special{language
"Scientific Word";type "GRAPHIC";maintain-aspect-ratio TRUE;display
"USEDEF";valid_file "F";width 5.0756in;height 3.0545in;depth
0pt;original-width 5.0211in;original-height 3.0104in;cropleft "0";croptop
"1";cropright "1";cropbottom "0";filename
'NoOfGammaTies.TIF';file-properties "XNPEU";}}

\section{\label{sec-discuss}\textbf{Discussion\protect\bigskip }}

\subsection{Limitations of Decision Making Factors}

The possibility of a suicide attack on a particular target is dependant not
only on the characteristics of the target---its symbolic or strategic value
and its vulnerabilities--- but also on the ambition, capabilities, and
sensitivities of the relevant terrorist organizations. This model as, based
on a generic analysis; in order to be truly effective, it would require far
more parameters than those shown---an analysis that is beyond the scope of
this paper. We have attempted to establish criteria for judging which attack
scenarios might prove \textquotedblleft attractive\textquotedblright\ to the
terrorist groups that might be interested in attacking the venue in question.

The value of the model discussed here lies not in its ability to reveal new
information, but rather in its usefulness as a planning and decision making
aid: a means to visualize data so that trends may be readily apparent where
they might otherwise have been lost in a heap of data. Naturally, the
resulting numbers are only as good as the information that goes into the
model. In effect, the model should be viewed as merely a template for better
organizing our knowledge; without that knowledge, the template is empty of
all content. As our judgments regarding the affinities and sensitivities of
these groups are necessarily approximate and since these input parameters
are associated with significant uncertainties, the resulting risk assessment
is also subject to large uncertainties due to error propagation. Thus when
based on reliable information, the model is a very useful tool for knowledge
management, data visualization and risk assessment.

\subsection{Possible Future Enhancements}

Following directions for future research may be identified:

\begin{enumerate}
\item This method can be further enhanced to include data-mining abilities.
This raw data would provide the basis for a continual reevaluation of the
organizational factors of interest that enter into the equation. The result
will be a dynamic threat assessment apparatus which could be programmed to
\textquotedblleft flag\textquotedblright\ certain types of venues and cities
when the relevant threat indices reach a critical level. The end product
would be applicable to governments, installation security, public transport
officials, and a host of other sectors that could be targeted by terrorists.

\item The algorithm in its present form is of order  is of following order $%
O\left( 2^{n}\right) $ of complexity as for $n$ cities in $X$ computer has
to carry out exactly $2^{n}$ comparisons due to first loop i.e.%
\begin{equation*}
FirstLoop\left( comparisons\right) +SecondLoop\left( additions\right)
=2^{n}+n.
\end{equation*}%
In future we shall endeavour to optimize complexity-efficiency profile of
FSP. 

\item Representation of knowledge in the form of a discernibility matrix has
been found of great advantages, in particular it enables simple computation
of the core, reducts and other concepts in Rough Set Theory \cite%
{pawlak91bk,skowron91}. Our computation of $\rho \left( \psi _{i},\psi
_{j}\right) $ and $\chi \left( \psi _{i},\psi _{j}\right) $ closely resemble
a discernibility matrix. Research efforts may fruitfully be employed to
exploit this resemblance.
\end{enumerate}

\begin{conclusion}
For a given collection of cities we present a model for predicting
occurrence of suicide attacks. The model has been developed in two parts.
Factors which affect the prediction oriented decision making have been
considered in first part. Considering incomplete information and the use of
linguistic terms by experts, as two characteristic parts of the prediction
problem we use the Theory of Fuzzy Soft Sets. The theory is still in its
embryonic stage and has a rich potential to solve such typically ill-posed
decision problems. Second part of the model is an algorithm vz. FSP which
takes the assessment of factors given in Part 1 as its input and produces a
possibility profile of cities likely to receive the accident. Algorithm has
been illustrated by an example solved in detail. Example considers the case
of Pakistan's five big cities and their relevant factors for prediction
making. Simulation results for the algorithm have been presented which give
insight into the strengths and weaknesses of FSP.

Three different decision making measures have been simulated and compared in
our discussion. We also point a few possible directions for future work in
the direction of this paper. Our simulation results show that decision
measure $\Gamma _{1}$ has far less bias as compared to other measures.
Though there can be scenarios where the other two measures are found more
useful.
\end{conclusion}

\end{document}